\documentclass[sn-basic,iicol]{sn-jnl}% Basic Springer Nature Reference Style/Chemistry Reference Style with double column layout

\usepackage{graphicx}%
\usepackage{multirow}%
\usepackage{amsmath,amssymb,amsfonts}%
\usepackage{amsthm}%
\usepackage{mathrsfs}%
\usepackage[title]{appendix}%
\usepackage{xcolor}%
\usepackage{textcomp}%
\usepackage{manyfoot}%
\usepackage{booktabs}%
\usepackage{algorithm}%
\usepackage{algorithmicx}%
\usepackage{algpseudocode}%
\usepackage{listings}%
\usepackage{soul}
\usepackage{diagbox}
\usepackage{ulem}
\usepackage{breakurl}

\theoremstyle{thmstyleone}%
\theoremstyle{thmstyletwo}%

\theoremstyle{thmstylethree}%

\begin{document}

\include{bst}
\graphicspath{{figures/}{pictures/}{images/}{./}} % where to search for the images

\title[Article Title]{LiVisSfM: Accurate and Robust Structure-from-Motion with LiDAR and Visual Cues}

\author[1]{\fnm{Hanqing} \sur{Jiang}}\email{jianghanqing@sensetime.com}
\equalcont{These authors contributed equally to this work.}

\author[1]{\fnm{Liyang} \sur{Zhou}}\email{zhouliyang@sensetime.com}
\equalcont{These authors contributed equally to this work.}

\author[1]{\fnm{Zhuang} \sur{Zhang}}\email{zhangzhuang@sensetime.com}
\equalcont{These authors contributed equally to this work.} 

\author[1]{\fnm{Yihao} \sur{Yu}}\email{yuyihao@sensetime.com}

\author*[2]{\fnm{Guofeng} \sur{Zhang}}\email{zhangguofeng@zju.edu.cn}

\affil[1]{\orgdiv{SenseTime Research}, \orgname{SenseTime Group Inc.}, \orgaddress{\street{29F, Tianren Tower, No. 188, Liyi Road}, \city{Hangzhou}, \postcode{311215}, %\state{Zhejiang},
\country{China}}}

\affil*[2]{\orgdiv{State Key Lab of CAD\&CG}, \orgname{Zhejiang University}, \orgaddress{\street{No. 866, Yuhangtang Road}, \city{Hangzhou}, \postcode{310058}, %\state{Zhejiang},
\country{China}}}


\abstract{This paper presents an accurate and robust Structure-from-Motion (SfM) pipeline named LiVisSfM, which is an SfM-based reconstruction system that fully combines LiDAR and visual cues. Unlike most existing LiDAR-inertial odometry (LIO) and LiDAR-inertial-visual odometry (LIVO) methods relying heavily on LiDAR registration coupled with Inertial Measurement Unit (IMU), we propose a LiDAR-visual SfM method which innovatively carries out LiDAR frame registration to LiDAR voxel map in a Point-to-Gaussian residual metrics, combined with a LiDAR-visual BA and explicit loop closure in a bundle optimization way to achieve accurate and robust LiDAR pose estimation without dependence on IMU incorporation. Besides, we propose an incremental voxel updating strategy for efficient voxel map updating during the process of LiDAR frame registration and LiDAR-visual BA optimization. Experiments demonstrate the superior effectiveness of our LiVisSfM framework over state-of-the-art LIO and LIVO works on more accurate and robust LiDAR pose recovery and dense point cloud reconstruction of both public KITTI benchmark and a variety of self-captured dataset.}


\maketitle

\section{Introduction}
\label{sec:introduction}
Large-scale 3D reconstruction has already attracted lots of attentions due to its wide uses across various applications such as 3D navigation, mixed reality (MR), 3D simulation and digital twin cities, and continues to cause increasing interests recently due to the rising of differentiable neural 3D representations typified by Neural Radiance Fields (NeRF)~\cite{mildenhall2021nerf} and 3D Gaussian Splatting (3DGS)~\cite{kerbl20233d} technologies. With the population of digital cameras and Unmanned Aerial Vehicle (UAV), it is becoming more and more convenient to capture high-resolution multi-view photos or videos, which makes it an important research topic in computer vision and photogrammetry to reconstruct high-quality 3D point clouds or 3DGS models of natural scenes by carrying out Multi-View Stereo (MVS) or Gaussian Splatting on the captured imageries. Nevertheless, the captured scenes often contain textureless or non-lambertian surfaces, which bring great difficulties for MVS or 3DGS methods to achieve both geometric completeness and accuracy due to the visual ambiguity problem.

\begin{figure*}[htb!]
\centering
\includegraphics[width=0.8\linewidth]{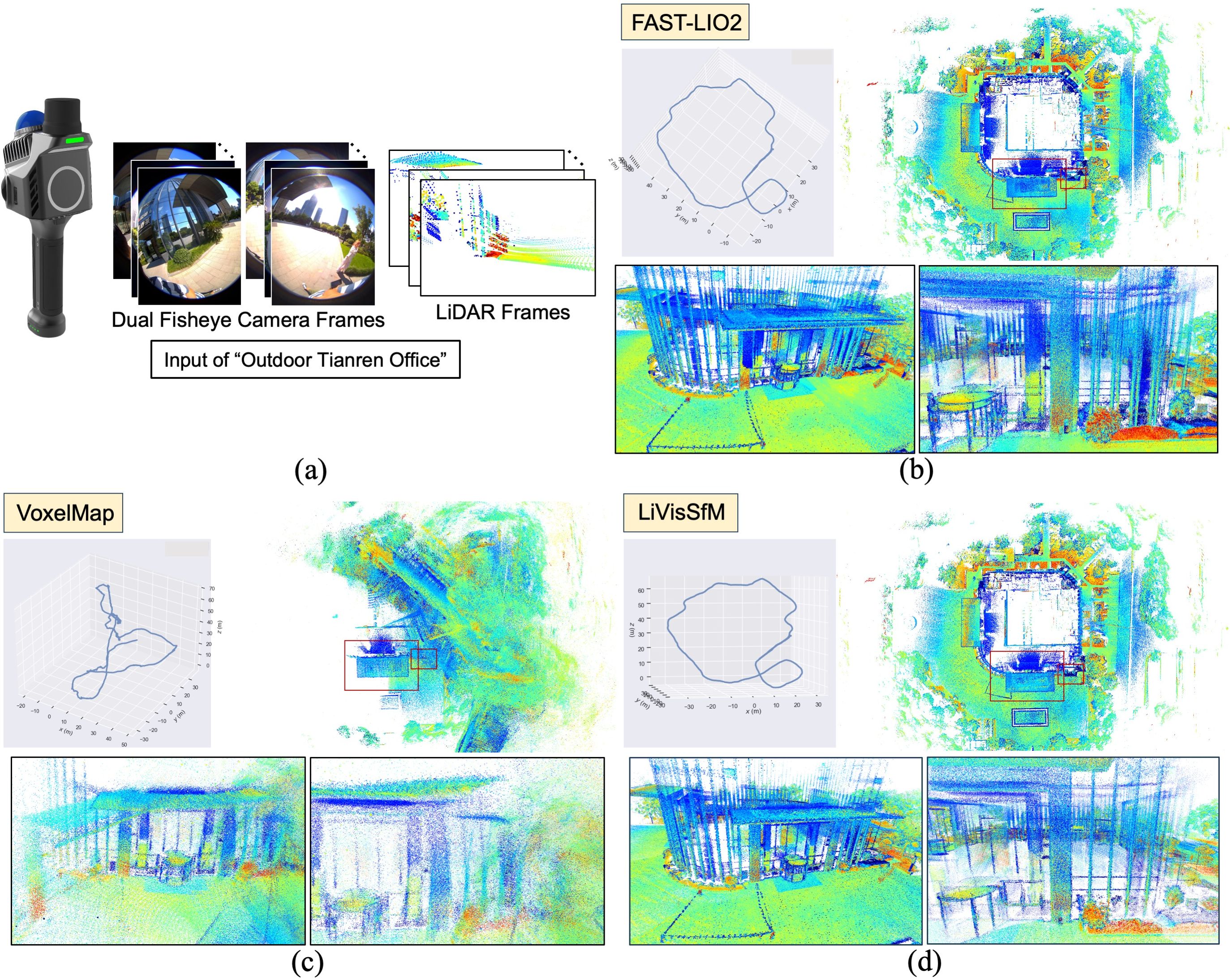}
\caption{3D point cloud reconstruction of ``Outdoor Tianren Office'' captured by MetaCam-Air handheld LiDAR scanner. (a) is the input LiDAR frames and fisheye camera frames. (b) is the estimated LiDAR trajectory and the fused LiDAR point cloud reconstructed by FAST-LIO2~\cite{xu2022fast}. (c) is the LiDAR trajectory and LiDAR point cloud reconstructed by VoxelMap~\cite{yuan2022efficient}, which collapses seriously. (d) is the LiDAR trajectory and LiDAR point cloud by our LiVisSfM pipeline, with the magnified regions highlighted in red rectangles to observe the local reconstruction details. It can be seen that FAST-LIO2 results in misalignment on the reconstructed dense point cloud, and our method can significantly eliminate long-range accumulation errors to produce finer local geometric details.}
\label{fig:teaser}
\end{figure*}

Recently, with the maturity of commercial LiDAR scanners, active LiDAR sensing technology is utilized by many LiDAR-inertial odometry (LIO) works such as ~\cite{xu2021fast,xu2022fast,yuan2022efficient,lin2022r}, to better solve the visual ambiguity introduced by the challenging weakly textured regions. These works usually align multiple LiDAR frames into 3D space and fuse the registered LiDAR points to a global point cloud without combination of visual tasks like feature matching, in order to alleviate the visual ambiguity problem. However, scenes with repeatable geometry like planar, spherical or cylinderical structures possibly cause misalignement of LiDAR frames due to geometric ambiguity. Besides, the frequently occurring reflective surfaces still lead to geometric inaccuracies in the scanned LiDAR data. Moreover, LiDAR point cloud from commercial scanners are commonly sparse in long distances with too wide gaps to capture sufficient high-frequency visual cues. These problems will certainly degrades the final reconstruction quality, and depend on additional visual cues captured by images and videos to be better solved. Some other LiDAR-inertial-visual odometry (LIVO) works like~\cite{lin2022r, zheng2022fast} try to loosely combine visual cues by coupling a visual-inertial odometry (VIO) subsystem with the main LIO system, where multi-view images are aligned with LiDAR map points by photometric error minimization for LiDAR point cloud colorization. Unfortunately, visual cues are seldom made full use of by these methods. Most of them only use photometric cues to guide image pose alignment for better point cloud colorization, while the point cloud fusion is performed only on LiDAR data, which makes it difficult for these methods to achieve high-quality large-scale 3D reconstruction on the long-range input LiDAR data. Large-scale scenes with long-range loops also bring great challenges to these LIO and LIVO works to eliminate accumulation errors or drifts, since they usually have no loop closure for LiDAR frames, and traditional visual loop closure might fail to handle complicated loops without enough feature matches to ensure sufficient loop edge connections. To ensure high-quality large-scale reconstruction with both completeness and accuracy, this paper proposes a novel LiDAR-visual Structure-from-Motion (SfM) pipeline named LiVisSfM, which is an SfM-based reconstruction system that fully combines LiDAR point clouds from active LiDAR sensor and visual cues from fish-eye digital cameras to recover accurate dense point cloud of challenging long-range scenes.

Fusion of LiDAR and visual cues for large-scale SfM is a less-researched topic. Most existing LIO and LIVO methods like~\cite{zhang2014loam, shan2018lego, xu2021fast, xu2022fast} utilize Kalman filtering for Inertial Measurement Unit (IMU) data fusion to enhance robustness of LiDAR frame tracking, with no use of visual information or visual cues used mainly for LiDAR colorization, which makes these methods rely heavily on both the quality of LiDAR data and the accuracy of IMU. Another problem of these methods is that the filtering-based systems are easy to be stuck into long-term drifts without explicit global BA of both LiDAR and visual maps. To better solve these problems, our offline LiDAR-visual SfM system fully integrates LiDAR and visual cues for LiDAR and camera frame registration, combined with a global Bundle Adjustment (BA) and loop closure for refinement of both LiDAR and camera frames, to get rid of dependence on IMU for robustness.

Our LiDAR-visual SfM applies a voxel map based LiDAR-visual pose estimation strategy to accurately register fish-eye camera frames and LiDAR point clouds to the global visual and LiDAR maps in an alternative way, with each LiDAR frame registered by maximizing the probabilities of new LiDAR points in the LiDAR voxel map based on our Point-to-Gaussian residual metrics. Meanwhile, a global BA is performed for optimizing LiDAR and camera poses together with LiDAR map and visual feature map in a time efficient way, combined with an explicit LiDAR and visual loop closure to gradually eliminate accumulated pose errors. After we have the optimized global SfM poses for both LiDAR and camera frames, each LiDAR point is projected to a time-closest camera frame to get its visual color. The colorized multi-frame LiDAR points are fused to a final complete global point cloud with accurate geometric details.
A large-scale example of ``Outdoor Tianren Office'' is shown in Fig.~\ref{fig:teaser} to demonstrate the better performance of our LiVisSfM framework in reconstructing a more complete dense point cloud with fewer accumulation drifts and more accurate geometric detail recovery, especially in textureless and reflective regions, compared to the state-of-the-art (SOTA) LIO and LIVO methods like FAST-LIO2~\cite{xu2022fast}, VoxelMap~\cite{yuan2022efficient} and LOAM~\cite{zhang2014loam}.

In summary, our LiVisSfM system contributes in the following main aspects:
\begin{itemize}
\item We propose a voxel map based LiDAR pose estimation approach which robustly registers LiDAR points in a Point-to-Gaussian residual metrics to solve the misalignment problem introduced by long-range accumulation errors.
\item A novel global LiDAR-visual BA scheme is proposed that jointly optimizes LiDAR and visual camera poses and updates voxel map in a time efficient way by adopting an incremental updating strategy. Meanwhile, explicit LiDAR and visual loops are established for pose graph optimization
to further eliminate accumulation errors.
\item Alternative estimation of LiDAR and visual poses and global LiDAR-visual bundle optimization are fully combined into an innovative LiDAR-visual SfM framework for accurate LiDAR and visual pose recovery and point fusion to get a high-quality dense point cloud of the scene without dependence on IMU.
\end{itemize}

This paper is organized as follows. Section \ref{sec:related-work} briefly presents related work. Section \ref{sec:system-overview} gives an overview of the proposed LiVisSfM system. The alternative LiDAR-visual pose registration and global LiDAR-visual bundle optimization are described in sections \ref{sec:visual-lidar-pose-estimation} and \ref{sec:lidar-visual-BO} respectively.
Finally, we give both qualitative and quantitative evaluations of our LiVisSfM pipeline in section \ref{sec:experiments}.

\begin{figure*}[htb!]
\centering
\includegraphics[width=0.95\linewidth]{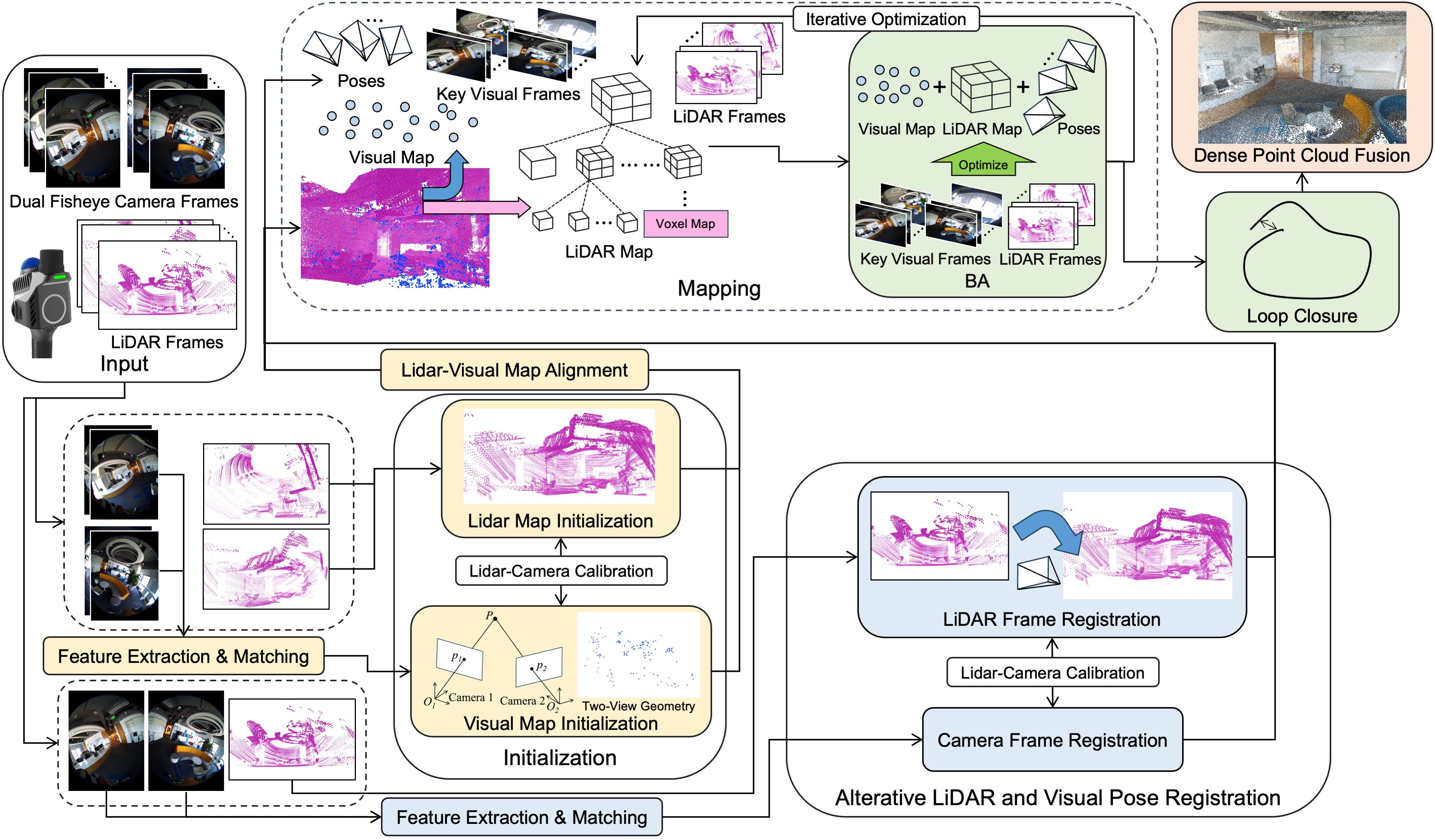}
\caption{System overview of our LiVisSfM, which consists of three main modules: LiDAR and visual map initialization, alternative LiDAR and visual pose registration, and mapping. The mapping module includes a global LiDAR-visual BA and an explicit loop closure carried out on a visual map with sparse feature points and a LiDAR map represented in voxel map structure. The multiple LiDAR frames are finally fused by the optimized LiDAR poses and colorized by the visual frames to a complete dense point cloud.}
\label{fig:system-overview}
\end{figure*}

\section{Related Work}
\label{sec:related-work}

\subsection{LiDAR-Inertial Odometry}
LiDAR-inertial odometry (LIO) is a filtering-based Simultaneous Localization and Mapping (SLAM) framework fusing LiDAR point clouds and IMU data. LOAM~\cite{zhang2014loam} is a foundational LIO work, which consists of three main module: LiDAR feature extraction, odometry and mapping. Subsequent works such as Lego-LOAM~\cite{shan2018lego} and LIO-SAM~\cite{shan2020lio} have similar system structures. Lego-LOAM~\cite{shan2018lego} is a loosely-coupled system with the IMU measurements not used in the optimization step. LIO-SAM~\cite{shan2020lio} is a tightly-coupled system of LiDAR and IMU with a sliding window of LiDAR keyframes introduced to reduce the computational complexity. Both works use KD-Tree representation for point cloud map, which is not efficient enough for large-scale scenes. To overcome this shortcoming, FAST-LIO~\cite{xu2021fast} introduces an incremental KD-Tree structure to organize point cloud map, which however also leads to increased complexity in map management. VoxelMap~\cite{yuan2022efficient} proposed an efficient probabilistic adaptive voxel mapping method as an alternative to point cloud map. VoxelMap++~\cite{wu2023voxelmap++} further improves upon this by designing a plane merging module using the union-find algorithm, which conserves resources and improves plane fitting accuracy. Most of these LIO methods require high-precision IMU whose state degrades easily in environments with similar structures, or relies on uniform motion model assumption to be optimized. Besides, LiDAR map management is time consuming for these methods, and usually requires complex data structures to reduce complexity. In comparison, our LiDAR-visual SfM utilizes an incremental updating strategy for highly efficient voxel map management, and combines LiDAR and visual cues for robust pose estimation without dependence on IMU.

\subsection{LiDAR-Inertial-Visual Odometry}
Multi-sensor fusion has been proven to be an effective way to achieve accurate and robust pose estimation, so that more and more LiDAR-visual and LiDAR-inertial-visual fusion frameworks have been proposed. VLOAM~\cite{zhang2018laser} uses a loosely coupled VIO to provide a motion prior for LiDAR odometry. Lic-Fusion~\cite{zuo2019lic} is a tightly coupled LiDAR-inertial-visual fusion framework, which uses the Multi-State Constrained Kalman Filter (MSCKF) framework to combine multi-source features. Lic-Fusion2~\cite{zuo2020lic} introduces a novel sliding-window plane-feature tracking algorithm for efficiently processing 3D LiDAR point cloud based on Lic-Fusion~\cite{zuo2019lic}. LVISAM~\cite{shan2021lvi} attempts to handle multi-source data of camera, LiDAR and IMU using a tightly coupled graph optimization based framework. R$^2$Live~\cite{lin2021r} propose parallel processing of the LIO and VIO modules. R$^3$Live~\cite{lin2022r} takes into account some camera photometric models based on the R$^2$Live method. FAST-LIVO~\cite{zheng2022fast} integrates LiDAR, camera and IMU data into the Error-State Iterated Kalman Filter (ESIKF), which can be updated via LiDAR and visual observations. Considering that LIO has better precision than VIO, SRLIVO~\cite{yuan2024sr} suggests using LIO for state estimation and employs a sweep reconstruction method for data synchronization. These LIVO methods usually require strict time synchronization between LiDAR and visual frames, and are prone to suffer accumulation error in long-range reconstruction scenarios, which can be effectively solved in our system by explicit LiDAR-visual loop closure.

\subsection{LiDAR-Visual Mapping and Reconstruction}
LiDAR-based mapping system easily degenerates in scenes with insignificant structural features, which can be enriched by visual cues. Visual SfM can generate textured reconstruction for large scale scene, which however also faces the challenges in low-texture areas and repetitive patterns. Shao et al.~\cite{shao2019stereo} couple VIO with LiDAR mapping and LiDAR enhanced visual loop closure. DV-LOAM~\cite{wang2021dv} first uses a two-staged visual odometry (VO) to get coarse poses, and then refines them by LiDAR mapping module. Amblard et al.~\cite{amblard2021lidar} propose to establish 2D-to-3D line correspondences between images and LiDAR frames to estimate LiDAR poses, which is then refined by line-based BA. The detected and optimized 3D lines are also used to improve the quality of final 3D geometry. PanoVLM~\cite{tu2023panovlm} uses a LiDAR-assisted global SfM to estimate the coarse camera and LiDAR poses, which are then refined by 2D-to-3D line correspondences between image and LiDAR frames respectively. Multi-view images with the refined poses are finally fused into a complete dense 3D map. In recent years, some novel view synthesis (NVS) based methods have also been used for LiDAR-visual reconstruction. SiLVR~\cite{tao2024silvr} present a neural-field-based large-scale reconstruction system that fuses LiDAR and visual data to generate high-quality reconstruction that is geometrically accurate and captures photo-realistic textures. Liv-GaussMap~\cite{hong2024liv} uses a LiDAR-inertial system with size-adaptive voxels to obtain initial poses for surface Gaussians which are then refined by photometric gradients to optimize the quality and density of LiDAR measurements. Its pipeline is compatible with various types of LiDAR of LiDAR-Visual method. These LiDAR mapping and reconstruction methods also require strict synchronization between LiDAR and camera sensors, while our method can relax this limitation up to a few deciseconds by our alternative LiDAR and visual frame registration.

\section{System Overview}
\label{sec:system-overview}

The user scans a large-scale scene with an commercial 3D scanner, which consists of a LiDAR sensor to capture a continuous sequence of LiDAR point clouds of the scene denoted as $\mathcal{L}$, with each LiDAR point cloud ${\mathcal{L}_t}$ called a LiDAR frame at timestamp $t$, and two fish-eye cameras to take synchronized multi-view video streams of the scene, which we denote as $\mathcal{I}_1$ and $\mathcal{I}_2$, without extra sensors like IMU. The LiDAR frames and the fish-eye camera frames are not strictly synchonized, up to hundreds of milliseconds. It is guaranteed here that the LiDAR sensor and the fish-eye cameras contain the factory intrinsic parameters and extrinsic parameters, based on which the LiDAR sensor and the fish-eye cameras are already calibrated with each other. Our LiVisSfM system is applied for the input LiDAR frames and multi-view fish-eye images to robustly reconstruct an accurate dense point cloud of the captured scene, which we denote as $\mathcal{P}$.

We carry out our LiDAR-visual SfM scheme to jointly estimate LiDAR and visual camera poses. Our system consists of three main modules: LiDAR and visual map initialization, alternative LiDAR and visual pose registration, and mapping. Firstly, feature extraction and matching are performed among the multi-view visual frames across the two fisheye cameras. Then, a visual map is initialized from the visual frames by two-view geometry of visual camera frames, where the poses of two LiDAR frames closest to the two registered visual frames by timestamps are calculated through the Iterative Closest Points (ICP) algorithm to build an initial LiDAR map represented in the form of voxel map~\cite{yuan2022efficient} organized by hash table, with each hash entry divided into an Octree, depending on the planarity of the probability distribution of the LiDAR points in the voxel. The initial LiDAR map is further aligned to the
visual map via Least Squares Method (LSM). After LiDAR and visual map initialization finishes, the rest visual camera and LiDAR poses are alternately estimated, with a pair of newly estimated synchronized camera poses provided as initial prior for pose estimation of a new LiDAR frame, by maximizing the probabilities of the new frame's LiDAR points located at corresponding voxels in the LiDAR map.
Meanwhile, a global LiDAR-visual BA is performed in the mapping module to jointly optimize LiDAR and visual camera poses in an efficient way,
with an explicit loop closure to further eliminate accumulated pose errors.
Both LiDAR-visual pose estimation and global LiDAR-visual BA are based on our Point-to-Gaussian residual metrics and incremental voxel updating strategy.
After we have the optimized global LiDAR and visual camera poses, the multi-frame LiDAR points are colorized by visual frames and fused together to a final complete global point cloud with accurate geometric details.
The proposed LiDAR-visual SfM scheme is outlined in Fig.~\ref{fig:system-overview}, with our method details introduced in the following sections.

\section{Alternate Estimation of Visual and LiDAR Poses}
\label{sec:visual-lidar-pose-estimation}
\begin{figure}[htp!]
\includegraphics[width=1.0\linewidth]{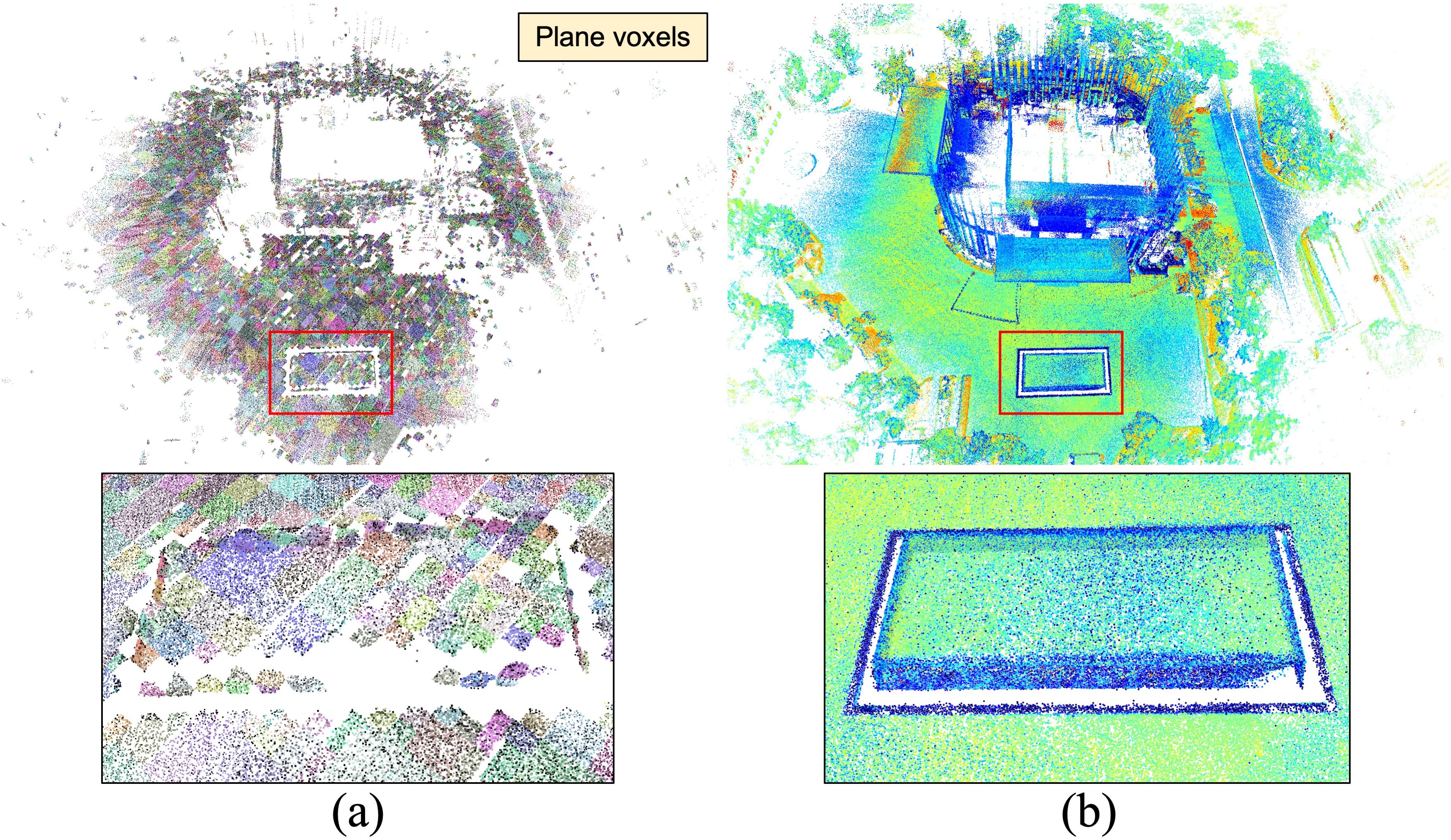}
\caption{Illustration of plane voxels in case ``Outdoor Tianren Office''. (a) Extracted plane voxels fusing all the registered LiDAR frames. (b) The finally fused dense point cloud, from which we can see that only the planar structures form the plane voxels, which no more LiDAR point will be inserted into.}
\label{fig:plane-voxels}
\end{figure}

\begin{figure*}[htb!]
\centering
\includegraphics[width=0.93\linewidth]{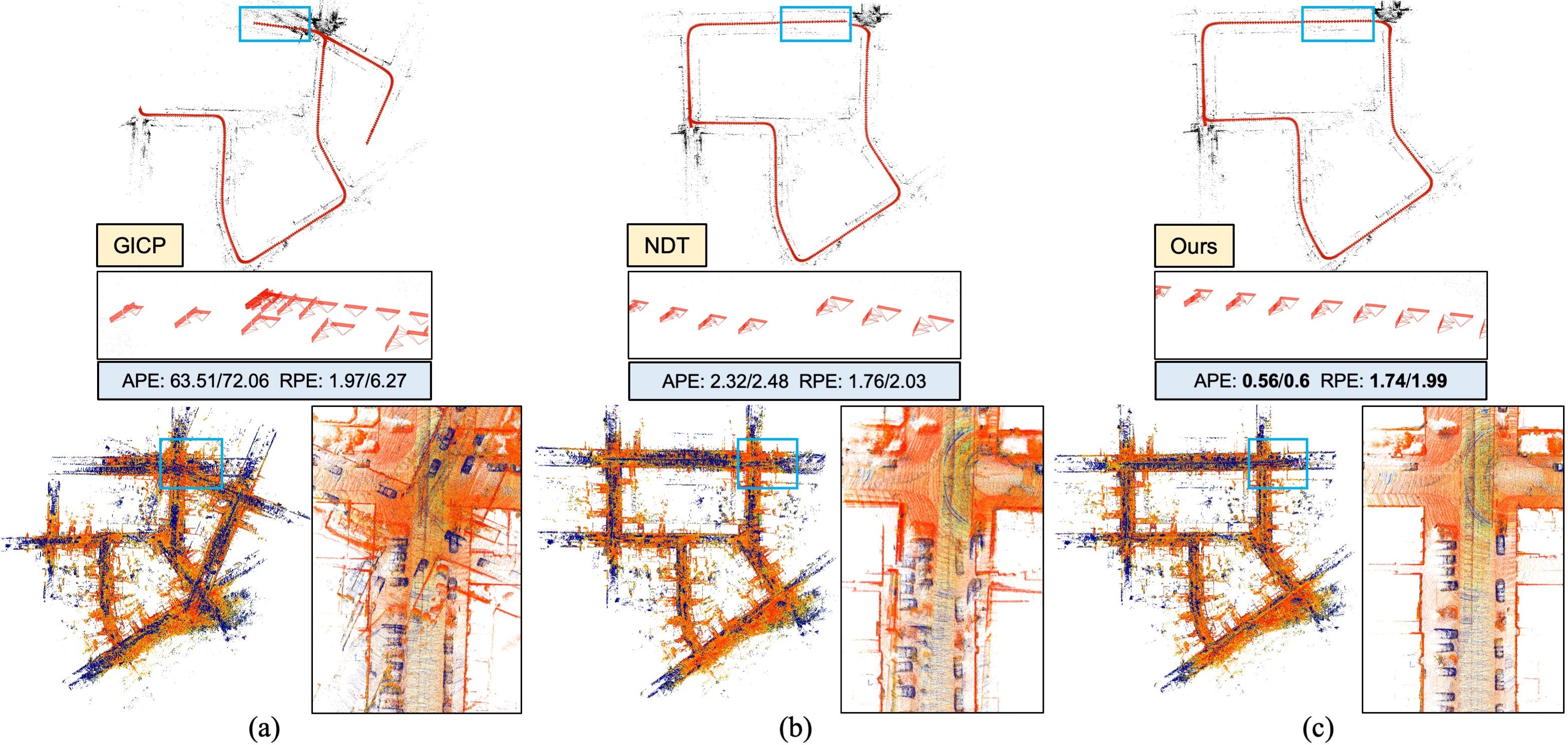}
\caption{Alternative estimation of LiDAR and visual poses. (a-c) are the estimated LiDAR poses of GICP~\cite{segal2009generalized}, NDT~\cite{biber2003normal} and our method respectively on KITTI ``07'' sequence, with LiDAR frames fused together to qualitatively compare the pose accuracies of the three methods, and the APE and RPE of the estimated LiDAR poses measured in MAE/RMSE in meters compared to the KITTI's GT poses given to quantitatively evaluate the pose accuracies. It can be seen that GICP has obvious pose estimation errors compared to NDT, but NDT still shows long-range accumulation error in the fused LiDAR point cloud. Our method performs the best in both the estimated LiDAR poses and the fused point cloud, which can also be verified by the highest accuracy in APE and RPE.}
\label{fig:pose-estimation}
\end{figure*}

LiDAR pose estimation module aligns each LiDAR frame to the LiDAR map in the global SfM coordinates. The traditional feature-based point cloud alignment approaches first extract geometric features such as points, lines and surfaces, and then perform LiDAR feature matching~\cite{zhang2014loam}, or use point-to-point ICP to align two point clouds like GICP~\cite{segal2009generalized}, or directly align the probability distributions of two point clouds as NDT~\cite{biber2003normal}. Feature-based or Point-to-Point methods like GICP are easily affected by LiDAR point uncertainty, while Point-to-Plane methods such as VoxelMap~\cite{yuan2022efficient} is prone to some under-constrained situations where there are no more than two orthogonal planes, which makes these methods sensitive to imperfect LiDAR data. The NDT method~\cite{biber2003normal} uniformly divides the scene, calculates the distribution of points separately in each divided region, and then estimates the LiDAR pose by maximizing the probabilities of LiDAR points in their located voxels. This strategy can achieve good registration for consistent LiDAR point clouds regardless of the uncertainties of LiDAR points, but is more susceptible to the inconsistent distributions of two LiDAR point clouds with a very close distance to each other, which can lead to mismatching or accumulated errors in complex environments.

To better handle large-scale LiDAR data in challenging scenes, we propose a novel alternate estimation approach for both visual camera and LiDAR poses. As in COLMAP~\cite{schoenberger2016sfm}, a latest visual frame is picked with the best feature matching score to the visual map to estimate its camera pose. A new LiDAR frame time-closest to the the latest visual camera frame is then picked as the next frame to register.
Here we enforce the selected LiDAR frame to be close to the previous LiDAR frames in both position and orientation, by checking whether its orientation difference from the nearest previously registered LiDAR frame is within ${60^{\circ}}$ to avoid challenging LiDAR alignment situations. It is worth noting that unlike LIO and LIVO works which perform online sequential registration in temporal domain, our offline frame selection strategy guides a more empirical order of LiDAR and visual frame registration that gradually expands from the previous registered set of frames, which does not have to strictly follow the time sequence of input frames.

For each newly selected LiDAR frame,
we use the estimated camera pose as the initial values for estimation of the LiDAR pose, which is adjusted by the relative extrinsics between fish-eye camera and LiDAR sensor. After that, the LiDAR points are fused to the global LiDAR map according to the estimated pose. In the fused LiDAR map, all LiDAR points are organized by voxel map division indexed by a hash table with size $3m \times 3m \times 3m$ for each hash entry,  
inside which is a set of LiDAR point Gaussian probability distributions organized by an Octree structure according to our proposed scheme combining both VoxelMap~\cite{yuan2022efficient} and NDT~\cite{biber2003normal} works. Here we make some improvements to the original NDT method. Specifically, we construct an Octree of LiDAR point Gaussian distributions for each hash entry in the global LiDAR voxel map.
When a new LiDAR frame is registered, its LiDAR points are fused into the corresponding hash entries they locate inside the voxel map, while the Gaussian distributions of LiDAR points inside these hash entries are updated, which is equivalent to change from the Frame-to-Frame scheme into Frame-to-Model of NDT method, similar to the ICP approach used in KinectFusion~\cite{izadi2011kinectfusion}. When sufficient LiDAR points contained in an Octree node can be successfully fit to a plane, the node division will stop to make a plane voxel leaf, into which no more LiDAR points will be inserted further, so as to effectively control the size of the entire voxel map. Otherwise, the node continues to be divided into leaf nodes, in which new probability distributions of the LiDAR points is determined.
To determine whether the current node becomes a plane voxel, we calculate the covariance matrix of the voxel node according to the LiDAR points located inside it and perform eigenvalue decomposition of the calculated covariance:
\begin{equation}
\begin{array}{*{20}{l}}
\mu_{\bf{v}} = \frac{1}{|\mathcal{L}({\bf{v}})|} \sum_{{\bf{x}}_p \in \mathcal{L}({\bf{v}})} {\bf{x}}_p, \\ [6pt]
\Sigma_{\bf{v}} = \frac{1}{|\mathcal{L}({\bf{v}})|} \sum_{{\bf{x}}_p \in \mathcal{L}({\bf{v}})}({\bf{x}}_p - \mu_{\bf{v}})^{\top}({\bf{x}}_p - \mu_{\bf{v}}), \\ [6pt]
{e}_{1}, {e}_{2}, {e}_{3} = EigenSolver(\Sigma_{\bf{v}}),
\end{array}
\label{eq:covariance-matrix}
\end{equation}
where ${\bf{v}}$ is the current voxel node in its voxel map Octree, ${\bf{x}}_p$ is a LiDAR point inside voxel ${\bf{v}}$, and $\mu_{\bf{v}}$ is the mean position of all the LiDAR points located inside ${\bf{v}}$ denoted by $\mathcal{L}({\bf{v}})$, and $\Sigma_{\bf{v}}$ is the covariance of the LiDAR points $\mathcal{L}({\bf{v}})$.
We perform eigenvalue decomposition to obtain three eigenvalues sorted from large to small, namely $e_1, e_2, e_3$. If the ratio of its minimum eigenvalue $e_3$ and its maximum one $e_1$ becomes extremely small, and the ratio between the second maximum eigenvalue $e_2$ and the maximum one $e_1$ is large enough,
namely ${e}_{3} / {e}_{1} < \sigma_d$ and ${e}_{2} / {e}_{1} > \sigma_s$, the node is considered as a plane voxel, which is the criterion we use for determining planarity of the current voxel. $\sigma_d$ and $\sigma_s$ are thresholds which we set to $0.03$ and $0.5$ respectively for our experiments. In this way, for Octree leaves with planarity, we can successfully construct plane voxels to represent planar structures. An example of the extracted plane voxels in case ``Outdoor Tianren Office'' is shown with various voxels represented in pseudo colors in Fig.~\ref{fig:plane-voxels}.

Our LiDAR pose estimation aims to maximize Frame-to-Model probability $D(\mathcal{L}_t)$ in the form of Point-to-Gaussian residual metrics as:
\begin{equation}
\begin{array}{*{20}{l}}
{\bf{x}}_w = \mathbf{M}_t^l {\bf{x}}_l, \\ [6pt]
{\bf{x}}_n = \vec{{\bf{n}}} ({\bf{x}}_w - \mu_{\bf{v}}), \\ [6pt]
D_v({\bf{x}}_l, {\bf{v}}) = w({\bf{x}}_l) (1 - \exp(-{{\bf{x}}_n}^{\top} \Sigma_{\bf{v}}^{-1} {\bf{x}}_n)), \\ [6pt]
D(\mathcal{L}_t) = 
\sum_{{\bf{x}}_l \in \mathcal{L}_t} {D_v({\bf{x}}_l, {\bf{v}})},
\end{array}
\label{eq:point-to-gaussian}
\end{equation}
where ${\bf{x}}_l$ is a LiDAR point in local coordinates of LiDAR frame $\mathcal{L}_t$ and ${\bf{x}}_w$ is the corresponding LiDAR point transformed to global coordinates by pose $\mathbf{M}_t^l = [\mathbf{R}_t^l | \mathbf{t}_t^l]$, with $\mathbf{R}_t^l$ and $\mathbf{t}_t^l$ the rotation matrix and translation vector of $\mathcal{L}_t$ to be estimated. ${\bf{v}}$ is the voxel which ${\bf{x}}_w$ locates inside, $\vec{{\bf{n}}}$ is the minimum eigen vector of $\Sigma_{\bf{v}}$, and ${\bf{x}}_n$ is the projection vector of ${\bf{x}}_p$ in $\vec{{\bf{n}}}$. $w({\bf{x}}_l)$ is a normal distribution weight about the LiDAR point intensity defined as $1 - \exp^{-\mathcal{U}({\bf{x}}_l)^2 / 100}$, with $\mathcal{U}({\bf{x}}_l)$ denoting the intensity of ${\bf{x}}_l$, whose value is normalized to ${\left[{0, 255}\right]}$. This Frame-to-Model probability definition uses an anisotropic Gaussian ellipsoid whose principal dimensions determined by the eigenvalues of the voxel node to better model the LiDAR point distribution of this voxel. It is worth noting that for both plane voxels and other Octree leaf nodes without planarity, we calculate the LiDAR point probabilities in the same way as Eq.~\ref{eq:point-to-gaussian}.

From the comparison of our LiDAR registration with two SOTA LiDAR registration methods GICP~\cite{segal2009generalized} and NDT~\cite{biber2003normal} on the KITTI ``07'' sequence shown in Fig.~\ref{fig:pose-estimation}, with the evaluated Absolute Percentage Error (APE) and Relative Percentage Error (RPE) of the estimated LiDAR poses measured in Mean Absolute Error (MAE) and Root Mean Squared Error (RMSE) compared to the KITTI's ground-truth (GT) LiDAR poses to quantitatively compare the pose accuracies. We can see from the comparison that our method achieves the best reconstruction in both LiDAR poses and the fused LiDAR point cloud, as also verified by the highest accuracy on APE and RPE, while both GICP and NDT generate certain pose misalignment or accumulation errors.

\section{Global LiDAR-visual Bundle Optimization}
\label{sec:lidar-visual-BO}
Long-range scene reconstruction inevitably introduces accumulative errors, which is often eliminated through global BA in traditional visual SfM. Different from original visual BA~\cite{triggs2000bundle}, we perform a global LiDAR-visual BA to jointly optimize all the LiDAR and visual camera poses together, which involves the updating of both the LiDAR voxel map and the visual map. Here we briefly introduce the original visual BA algorithm. Visual SfM uses feature tracks extracted from multi-view input images to reconstruct sparse map points while solving intrinstic and extrinsic parameters for each input image. For each image $ I_t^c \in \mathcal{I}_c$ with $c = 1, 2$ indexing the two fisheye cameras, we use $ \mathbf{K}_t^c $ and $ \mathbf{M}_t^c  = [ \mathbf{R}_t^c | \mathbf{t}_t^c ] $ to represent its intrinsic parameters and extrinsic pose respectively, with $\mathbf{R}_t^c$ and $\mathbf{t}_t^c$ being the rotation and translation respectively. All the intrinsic parameters, extrinsic poses and 3D feature map points are refined together in visual BA optimization by minimizing the following energy function:
\begin{equation}
E_\mathcal{I} = \sum_{\mathbf{x}_k \in \mathcal{X}} \sum_{\mathbf{k}_t^c \in \mathcal{T}(\mathbf{x}_k)}
 \left\| \pi( \mathbf{M}_t^c \mathbf{x}_k ) - \mathbf{k}_t^c \right\|^2,
\label{eq:ordinary-BA}
\end{equation}
with $\mathcal{X}$ denoting the set of 3D map points, $\mathbf{k}_t^c$ being a 2D feature in $I_t^c$ corresponding to a 3D map point $\mathbf{x}_k$ inside $\mathcal{X}$, and $\mathcal{T}(\mathbf{x}_k)$ is the 2D track of $\mathbf{x}_k$. $ \pi(x,y,z) = (f_u x / z + c_u,f_v y / z + c_v) $ is the projection function, in which $f_u$ and $f_v$ are the focal lengths in $uv$ directions of $I_t^c$, and $(c_u, c_v)$ is the optical center. For LiDAR-visual joint BA optimization, we also refine the poses of LiDAR frames together by additionally calculating the Point-to-Gaussian residuals of all the LiDAR frames in the voxel map by the same way as Eq.~\ref{eq:point-to-gaussian}, which are added to the original BA energy function for minimization as:
\begin{equation}
\begin{array}{*{20}{l}}
E_\mathcal{L} = \sum_{\mathcal{L}_t \in \mathcal{L}} D(\mathcal{L}_t), \\ [6pt]
E = E_\mathcal{I} + E_\mathcal{L}.
\end{array}
\label{eq:LiDAR-visual-BA}
\end{equation}

\subsection{Incremental Voxel Map Updating}
During each iterative optimization of LiDAR-visual BA, the voxel map remains unchanged.
When the LiDAR poses change after our LiDAR-visual BA, the voxel map of LiDAR points need to be updated also in time, which will cost huge time consumption that seriously degrades the runtime performance, especially when most LiDAR poses are changed by global BA. 
In order to better handle this problem, we propose an incremental updating strategy to update voxel map in a time-efficient way. When the pose of a LiDAR frame changes, the positions of its LiDAR points also change in the voxel map. When a LiDAR point ${\bf{x}}_l$ jumps from one voxel $\bf{v}$ to another one $\bf{v}^{\prime}$ due to the updated pose, we need to remove the LiDAR point from the old voxel $\bf{v}$ by its location ${\bf{x}}_w = {\mathbf{M}}_t^l {\bf{x}}_l$ according to the old pose ${\mathbf{M}}_t^l$, and add the LiDAR point to the new voxel $\bf{v}^{\prime}$ according to its new position $\hat{\bf{x}}_w = \hat{\mathbf{M}}_t^l {\bf{x}}_l$ by the updated pose $\hat{\mathbf{M}}_t^l$. For the changed voxel $\bf{v}$ with its old mean position $\mu_{\bf{v}}$ and covariance matrix $\Sigma_{\bf{v}}$, we need to recalculate its new mean position $\hat{\mu}_{\bf{v}}$ and covariance matrice $\hat{\Sigma}_{\bf{v}}$ by Eq.~\ref{eq:covariance-matrix}. Actually, we figure out that since only one LiDAR point is deleted, there is no need to recalculate the mean and covariance on all the remaining LiDAR points. Instead, we can incrementally update its mean and covariance in a simpler way as:
\begin{equation}
\begin{array}{*{20}{l}}
\hat{\mu}_{\bf{v}} = \frac{1}{|\mathcal{L}({\bf{v}})| - 1}(|\mathcal{L}({\bf{v}})| \mu_{\bf{v}} - {\bf{x}}_w), \\ [6pt]
\hat{{\Sigma}}_{\bf{v}}(i,j) = \frac{|\mathcal{L}({\bf{v}})| ({{\Sigma}_{\bf{v}}}(i,j) + \mu_{\bf{v}}(i) \mu_{\bf{v}}(j)) - {\bf{x}}_w(i) {\bf{x}}_w(j)}{|\mathcal{L}({\bf{v}})| - 1} \\ [6pt]
\quad \quad \quad \quad - \hat{\mu}_{\bf{v}}(i) \hat{\mu}_{\bf{v}}(j), \quad \quad \forall i, j \in \{0, 1, 2\}, \\ [6pt]
\mathcal{L}({\bf{v}}) = \mathcal{L}({\bf{v}}) \backslash {\bf{x}}_w,
\end{array}
\label{eq:mean-covariance-sub}
\end{equation}
where ${\bf{x}}_w(i)$ represents the $i$-th component of the 3D point ${\bf{x}}_w$, which is the same for $\mu_{\bf{v}}(i)$ and $\hat{\mu}_{\bf{v}}(i)$, and ${\Sigma}_{\bf{v}}(i,j)$ is the $i$-th row and $j$-th column of the covariance matrix ${\Sigma}_{\bf{v}}$.
Similarly, for ${\bf{v}}^{\prime}$ with its old mean $\mu_{\bf{v}^{\prime}}$ and covariance $\Sigma_{\bf{v}^{\prime}}$, we incrementally update its new mean $\hat{\mu}_{\bf{v}^{\prime}}$ and covariance $\hat{\Sigma}_{\bf{v}^{\prime}}$ by adding the new LiDAR point $\hat{\bf{x}}_w$ as:
\begin{equation}
\begin{array}{*{20}{l}}
\hat{\mu}_{\bf{v}^{\prime}} = \frac{1}{|\mathcal{L}({\bf{v}^{\prime}})| + 1}(|\mathcal{L}({\bf{v}^{\prime}})| \mu^{\prime}_{\bf{v}} + \hat{\bf{x}}_w), \\ [6pt]
\hat{{\Sigma}}_{\bf{v}^{\prime}}(i,j) = \frac{|\mathcal{L}({\bf{v}^{\prime}})| ({\Sigma}_{\bf{v}^{\prime}}(i,j) + \mu_{\bf{v}^{\prime}}(i) \mu_{\bf{v}^{\prime}}(j)) + \hat{\bf{x}}_w(i) \hat{\bf{x}}_w(j)} {|\mathcal{L}({\bf{v}^{\prime}})| + 1} \\ [6pt]
\quad \quad \quad \quad - \hat{\mu}_{\bf{v}^{\prime}}(i) \hat{\mu}_{\bf{v}^{\prime}}(j), \quad \quad \forall i, j \in \{0, 1, 2\}, \\ [6pt]
\mathcal{L}({\bf{v}^{\prime}}) = \mathcal{L}({\bf{v}^{\prime}}) \cup \hat{\bf{x}}_w.
\end{array}
\label{eq:mean-covariance-add}
\end{equation}
The detailed derivation of Eq.~\ref{eq:mean-covariance-sub} and \ref{eq:mean-covariance-add} are provided in the appendix~\ref{sec:appendix}.

Therefore, the probability distribution of ${\bf{x}}_l$ on a voxel $\bf{v}$ defined in Eq.~\ref{eq:point-to-gaussian} can be updated as:
\begin{equation}
\begin{array}{*{20}{l}}
\hat{\bf{x}}_n = \vec{{\bf{n}}} (\hat{\bf{x}}_w - \mu_{\bf{v}}), \\ [6pt]
\hat{D}_v({\bf{x}}_l, {\bf{v}}) = w({\bf{x}}_l) (1 - \exp(-{\hat{{\bf{x}}}_n}^{\top} \hat{\Sigma}_{\bf{v}}^{-1} \hat{{\bf{x}}}_n)).
\end{array}
\label{eq:update-point-to-gaussian}
\end{equation}

After the LiDAR and visual poses are refined with LiDAR-visual BA optimization by Eq.~\ref{eq:LiDAR-visual-BA}, the LiDAR voxel map is updated in the following way. For each LiDAR frame $\mathcal{L}_t$ whose pose changes from ${\mathbf{M}}_t^l$ to $\hat{{\mathbf{M}}}_t^l$, we first transform the local LiDAR point cloud by the old pose ${\mathbf{M}}_t^l$ to the global coordinates to get an old position ${\bf{x}}_w$ for each LiDAR point ${\bf{x}}_l \in \mathcal{L}_t$ for finding the old voxel the point locates in, delete the point from the voxel, and then update its mean and covariance by Eq.~\ref{eq:mean-covariance-sub}. After the deletion operations for all the changed LiDAR frames are finished, addition operations continue. This time, each changed local LiDAR point cloud $\mathcal{L}_t$ is transformed by the new pose $\hat{{\mathbf{M}}}_t^l$ to the global coordinates. For each LiDAR point ${\bf{x}}_l \in \mathcal{L}_t$, its new global position $\hat{{\bf{x}}}_w$ is used to find the new voxel it is located inside, add the point to the voxel, and update its mean and covariance by Eq.~\ref{eq:mean-covariance-add}. If all the LiDAR points of a voxel are deleted, this degraded voxel is removed from its Octree. For all the changed voxels, their mean positions and covariance matrices are updated efficiently for the incoming LiDAR frame registrations and the next time of LiDAR-visual BA. In this incremental updating way, the status of voxel map can be updated in a low computational cost even when many of the LiDAR poses are changed, regardless of the number of the LiDAR points contained in the voxels to be updated.

It is worth noting that for our implementation of LiDAR frame registration in section~\ref{sec:visual-lidar-pose-estimation}, our voxel map is gradually updated in the same incremental way as more and more new LiDAR frames are registered. When a new LiDAR frame $\mathcal{L}_t$ is registered by its estimated pose ${\mathbf{M}}_t^l$, each LiDAR point ${\bf{x}}_l \in \mathcal{L}_t$ is transformed to its global position ${\bf{x}}_w = {\mathbf{M}}_t^l {\bf{x}}_l$ to find the voxel it is added into, whose mean and covariance is then updated by Eq.~\ref{eq:mean-covariance-add}.

In fact, we also consider the influence of sparse visual map points in Eq.~\ref{eq:ordinary-BA} on the voxel map during the LiDAR-visual BA optimization.
Point-to-Gaussian residuals of the visual map points to the voxels in which they locates are minimized together with the Point-to-Gaussian residuals from the LiDAR points, so that a global optimization of the LiDAR map will also affect the refinement of the visual map, which in turn affects the visual camera pose refinement, to better achieve a joint bundled optimization of both LiDAR and visual camera poses.

In addition, we record the creation time of each voxel $\mathbf{v}$ as $T({\mathbf{v}})$ according to the LiDAR frame registration by which it is created for the first time, and the time of the current LiDAR frame ${\mathcal{L}}_t$ denoted by $T({\mathcal{L}}_t)$. The creation time is numbered by the LiDAR frame number starting from $0$ for the first LiDAR frame. When performing global LiDAR-visual BA, for all the participating voxels, we also add a time-related weight to the Point-to-Gaussian residuals to involve the time record of each voxel $\mathbf{v}$ by modifying Eq.~\ref{eq:update-point-to-gaussian} as:
\begin{equation}
\begin{array}{*{20}{l}}
\hat{D}_v({\bf{x}}_l, {\bf{v}}) = w_t({\mathbf{v}}) w({\bf{x}}_l) (1 - \exp(-{\hat{{\bf{x}}}_n}^{\top} \hat{\Sigma}_{\bf{v}}^{-1} \hat{{\bf{x}}}_n)), \\ [6pt]
\delta_t = |T({\mathcal{L}}_t) - T({\mathbf{v}})|, \\ [6pt]
w_t({\mathbf{v}}) = 38 / {\exp(\frac{\delta_t}{25} + 1)} + 1.
\end{array}
\label{eq:loop-closure-weight}
\end{equation}
Here we use the absolute difference between the current LiDAR frame number and the voxel creation time, because the voxel might be created with a time later than the time of current LiDAR frame according to our new frame selection strategy discussed in section \ref{sec:visual-lidar-pose-estimation}. According to Eq.~\ref{eq:loop-closure-weight}, when registering a newly coming LiDAR frame, if an earlier voxel is observed, we apply a larger weight to its residual. Here we bound $w_t({\mathbf{v}})$ up to $15$ through the sigmoid function. This strategy acts like ``an implicit loop closure'', which helps to further alleviate the accumulation of small errors, by considering more about the probability distributions of those voxels with time far apart than the time-close ones.

\subsection{Explicit LiDAR-visual Loop Closure}
\begin{figure*}[htb!]
\centering
\includegraphics[width=1.0\linewidth]{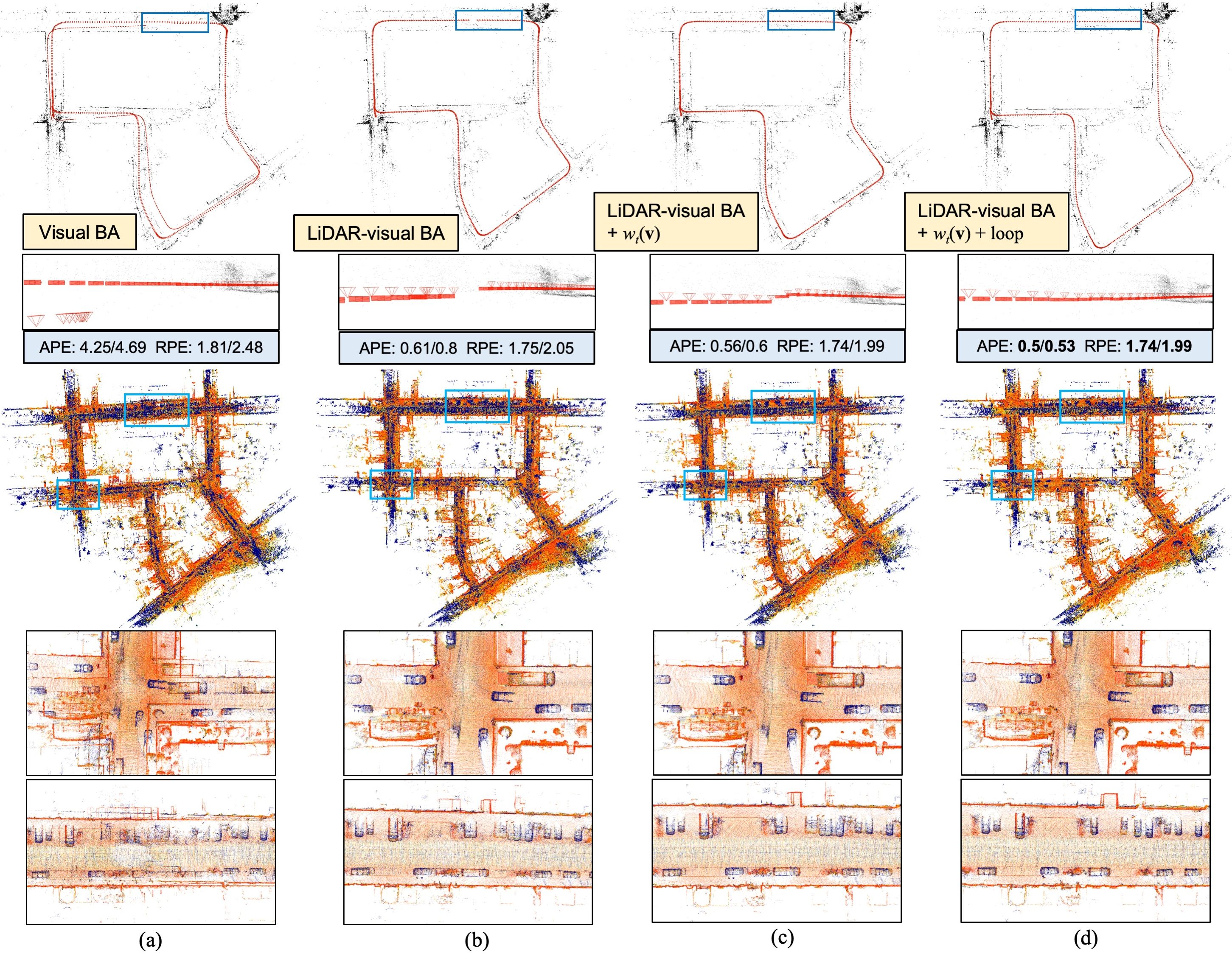}
\caption{Ablation study about LiDAR-visual BA, time-related weight and explicit loop closure. (a) With only visual BA, the LiDAR pose has obvious accumulation error. (b) LiDAR-visual BA significantly improves LiDAR pose accuracy, but accumulation drift still occurs. (c) After involving the time-related weight, the pose accumulation error is further reduced. (d) The combination of LiDAR-visual BA, time-related weight and explicit loop closure achieves the best reconstruction result on both the estimated LiDAR poses and the fused LiDAR point cloud as shown in the magnified regions highlighted by the blue rectangles, also with the best pose accuracy evaluated by APE and RPE in MAE/RMSE in meters.}
\label{fig:ablation-study}
\end{figure*}

Furthermore, we add an explicit loop closure to handle globally accumulated drifts in large-scale long-range scenarios.
Traditional loop closure methods for LiDAR scans usually employ local keypoint detection and matching in combination with the Bag-of-Words (BoW) approach~\cite{steder2011place}, or combine a global point cloud matcher with a novel registration algorithm to determine loop candidates~\cite{vlaminck2018have}. However, detecting distinctive keypoints in scenes with highly repeatable structures remains a challenge within 3D point cloud analysis.
In comparison, we construct loop closures for visual and LiDAR frames respectively to eliminate the global drift among time-consecutive LiDAR frames and visual camera frames.

\begin{figure}[htp!]
\centering
\includegraphics[width=1.0\linewidth]{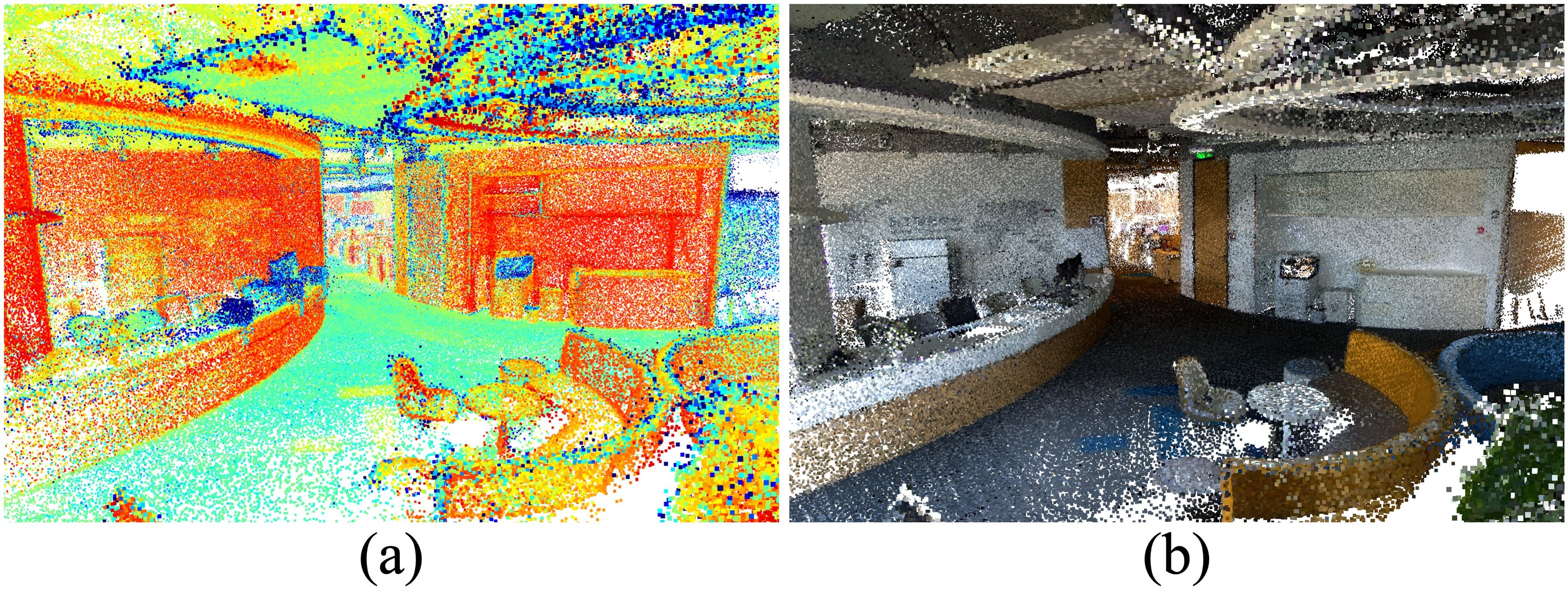}
\caption{LiDAR point cloud fusion of case ``Tearoom''. (a) The LiDAR point cloud fused by the optimized LiDAR poses, in pseudo color of LiDAR intensity. (b) The point cloud colorized by visual camera frames.}
\label{fig:dense_point_cloud}
\end{figure}

\begin{figure*}[htb!]
\centering
\includegraphics[width=1.0\linewidth]{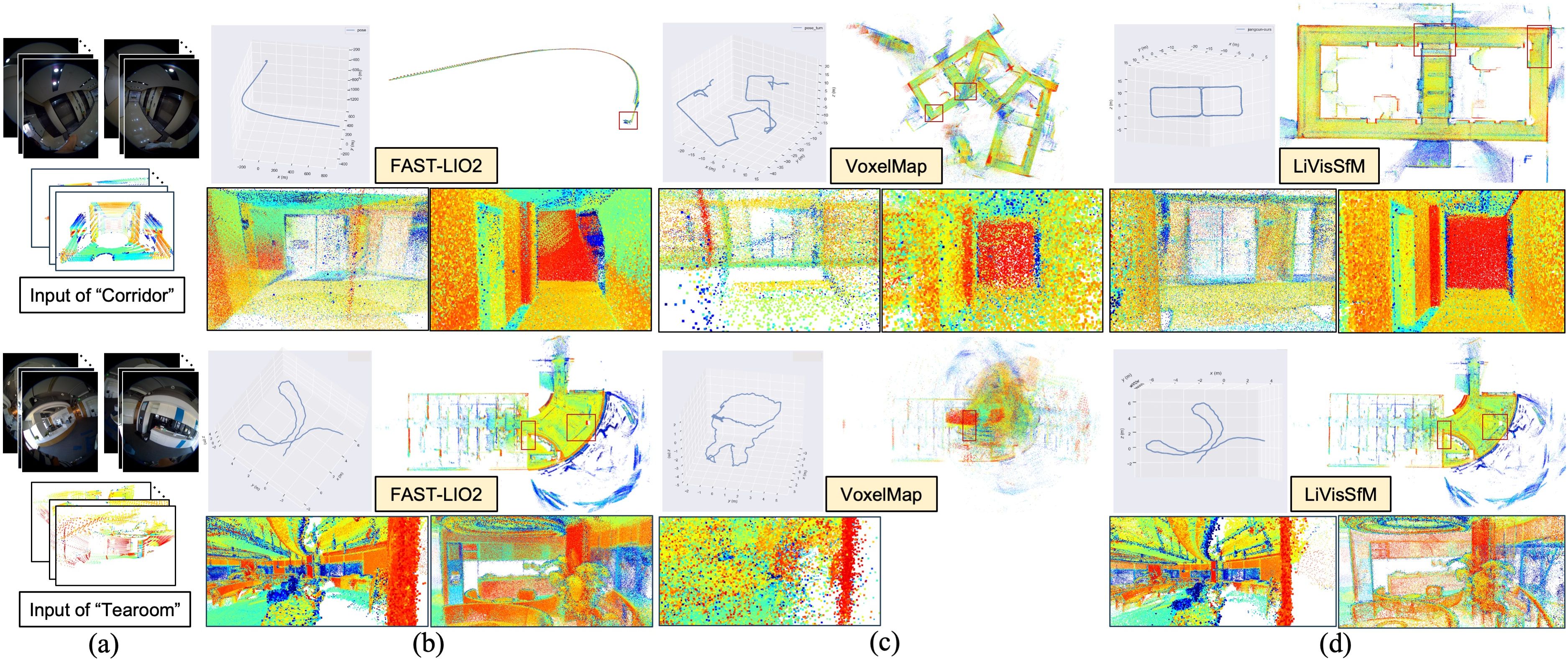}
\caption{(a) are the inputs of two self-captured cases ``Corridor'' and ``Tearoom''. (b-d) gives the reconstructed camera trajectories and fused dense point clouds of our LiVisSfM compared to FAST-LIO2 and VoxelMap on the two cases, with the magnified regions highlighted in red rectangles to show the reconstruction accuracy and robustness of our method.}
\label{fig:comparison-our-data}
\end{figure*}

\begin{figure*}[htb!]
\centering
\includegraphics[width=1.0\linewidth]{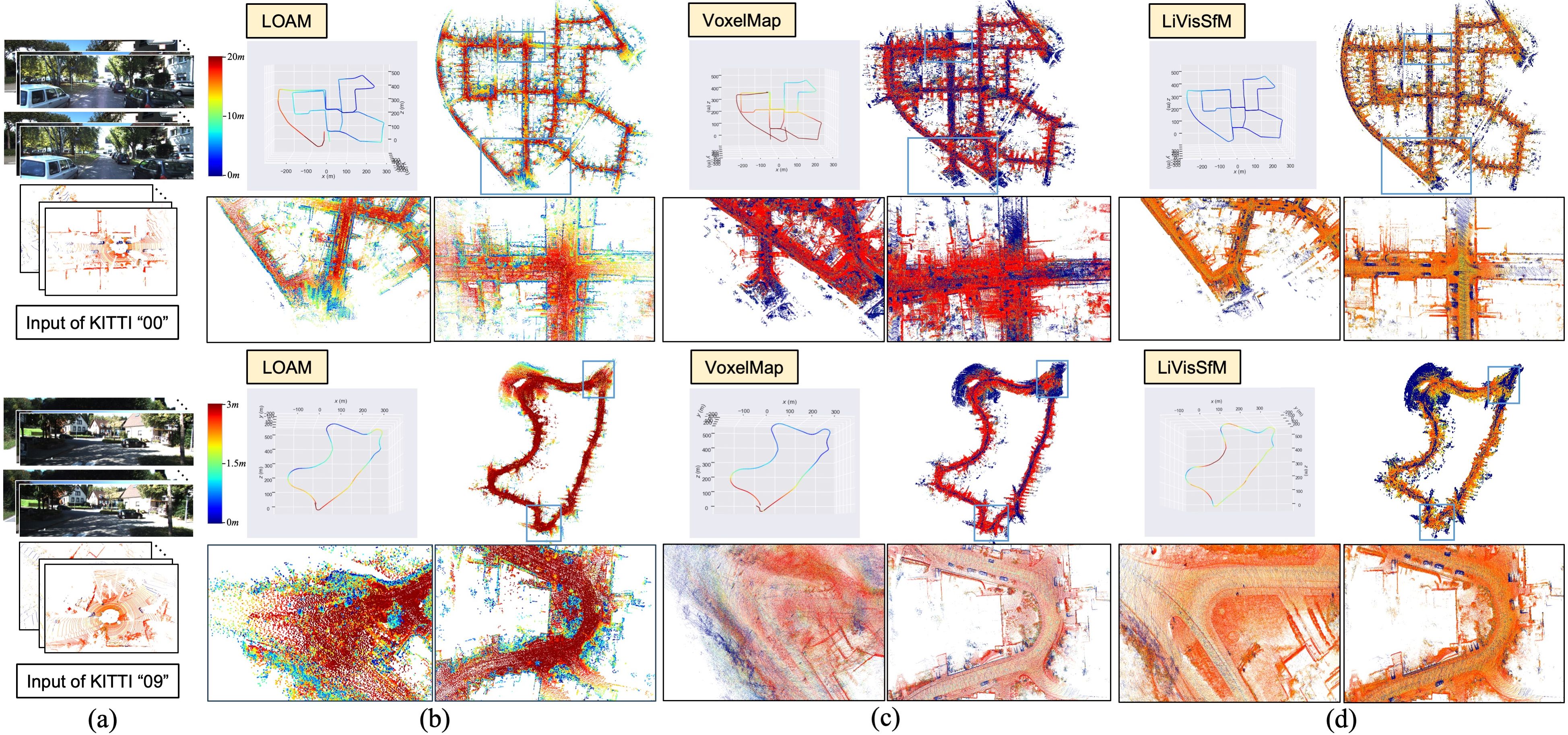}
\caption{(a) are the inputs of two KITTI sequences ``00'' and ``09''. (b-d) shows the reconstructed camera trajectories and fused dense point clouds of our pipeline compared to LOAM and VoxelMap on the two sequences, with the magnified blue rectangles to show the better accuracy of our LiDAR reconstruction with fewer drifts and accumulation errors.}
\label{fig:comparison-kitti}
\end{figure*}

Specially, we determine the potential drift among two time-consecutive fisheye camera frames based on the ratio between the number of consensus visual map points and the number of feature matches among the two frames, whose threshold is set as $0.1$ in the experiments. Meanwhile, the drifts among two time-consecutive LiDAR frames are also detected by the change in the rotation angle and the moving speed of LiDAR sensor. The change of the LiDAR moving speed is obtained by predicting the expected speed of the current two LiDAR frames from the previous three pairs of consecutive LiDAR frames through linear interpolation, and compute the ratio between the actual speed and the expected one, with the larger divided by the smaller. The LiDAR speed is computed by the distance between two consecutive LiDAR frames divided by their time interval. If the angle difference between the current two frames is greater than $\delta_\alpha$ or the speed ratio exceeds a thresold $\delta_s$, we consider that there is a LiDAR drift. The thresold is set to $\delta_\alpha = 3^{\circ}$ and $\delta_s = 2$ for KITTI sequences, and $\delta_\alpha = 10^{\circ}$ and $\delta_s = 3$ for our self-captured cases, considering that KITTI dataset are captured by driving platform that usually cruises smoothly, while our handheld capturing way is more shaking and jittering. After we detect all the drifts among visual camera and LiDAR frames, we carry out the explicit loop closure via pose graph optimization. Our pose graph is built on edges of visual frame pairs connected by top $5$ most matched visual features for each visual frame, edges of LiDAR frame pairs connected by top $5$ nearest positions, and loop closure edges from all the detected drifts among consecutive visual camera and LiDAR frames. For each drifted visual frame pair, we solve the Perspective-n-Point (PnP) problem for the image features on one visual frame and the sparse 3D points on the other frame to acquire their relative pose to construct the loop closure edge. For each drifted LiDAR frame pair, the relative pose of the camera frame pair time-closet to the two LiDAR frames are used as initial values to refine the relative pose in the global LiDAR map to construct the loop closure edge. Eventually, all the visual and LiDAR poses are globally adjusted through the pose graph optimization followed by a final time of LiDAR-visual BA to attain better global consistency.

We perform ablation experiments on our LiDAR-visual BA and loop closure, implicit and explicit loop closure to verify their effectiveness. Fig.~\ref{fig:ablation-study} shows the experimental results of the optimized LiDAR poses and the fused dense point clouds by Visual BA, LiDAR-visual BA, LiDAR-visual BA with time-related weight, and LiDAR-visual BA with both time-related weight and explicit loop closure on KITTI ``07'' sequence, with accuracy comparison of LiDAR trajectories also given by APE and RPE in MAE/RMSE. As can be seen from the comparison results, the recovered LiDAR trajectory of visual BA drifts easily. LiDAR-visual BA significantly improves the trajectory accuracy. When the time-related weight $w_t({\mathbf{v}})$ is added, the accumulation drift is further reduced. Finally, adding the explicit loop closure module helps to achieve the best LiDAR pose accuracy.

\subsection{Dense Point Cloud Fusion}
\label{sec:point-cloud-fusion}
After we have an optimized high-quality pose $\hat{\mathbf{M}}_t^l$ for each LiDAR frame $\mathcal{L}_t$, each LiDAR point ${\bf{x}}_l \in \mathcal{L}_t$ is transformed to their global position ${\bf{x}}_w = \hat{\mathbf{M}}_t^l {\bf{x}}_l$ to get a set of 3D points. Subsequently, each 3D point is projected
to the time-closest fisheye camera frame to get its visual color. Finally, the colorized 3D points of all LiDAR frames are combined to a dense point cloud we denoted as $\mathcal{P}$. An exemplar fused LiDAR point cloud of case ``Tearoom'' is shown in Fig.~\ref{fig:dense_point_cloud}.

\section{Experiments}
\label{sec:experiments}
In this section, we perform qualitative and quantitative evaluation of our LiVisSfM pipeline, whose core algorithms are implemented in C++, on various indoor and outdoor experimental cases including three KITTI autonomous driving sequences~\cite{geiger2012we} and four cases captured ourselves using MetaCam-Air handheld LiDAR scanner produced by Skyland Innovation. Each self-captured case includes two fisheye cameras and a non-repetitive LiDAR sensor, with each fisheye camera recording images with $3040 \times 4032$ resolution and synchronized to the other camera in $4$ Hz, and the LiDAR sensor scanning LiDAR frames with $200,000/s$ density in $10$ Hz. We first exhibit the reconstruction results on both the KITTI dataset~\cite{geiger2012we} and the self-captured cases for qualitative comparison of our work to other SOTA LIO and LIVO methods including FAST-LIO2~\cite{xu2022fast}, VoxelMap~\cite{yuan2022efficient} and LOAM~\cite{zhang2014loam}.
Then the quantitative comparison of our approach to the SOTA works are also given to evaluate accuracies of the recovered LiDAR trajectories and the fused dense LiDAR point clouds of the self-captured case ``ZJU-CCE'' on both MAE and RMSE with its GT model pre-captured by laser scanning device. We also report the time consumption on the stages of LiDAR-visual SfM for various cases to show the time efficiency of our method.

\subsection{Qualitative and Quantitative Evaluations}
\begin{figure*}[htb!]
\includegraphics[width=1.0\linewidth]{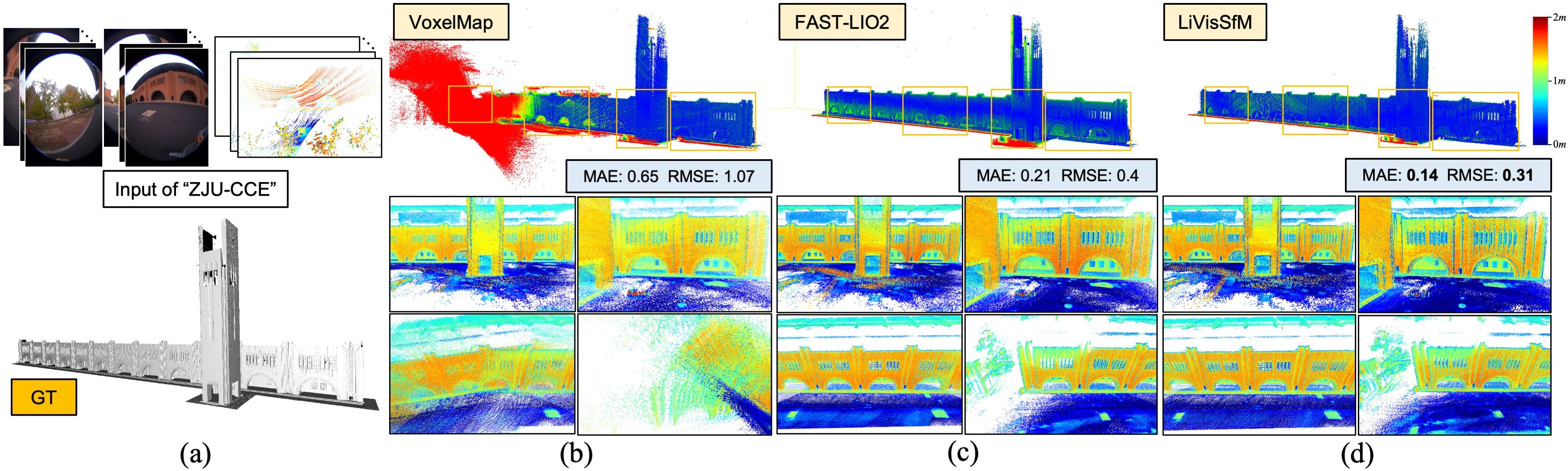}
\caption{Dense point cloud reconstruction accuracy of VoxelMap, FAST-LIO2 and our LiVisSfM on case ``ZJU-CCE'' compared with its GT model, with MAE and RMSE measurements in meters. (a) is the input of ``ZJU-CCE'' and the GT model. (b) shows the reconstructed LiDAR point cloud of VoxelMap, with obvious collapse on the left part of the building. (c) and (d) are the reconstruction results of FAST-LIO2 and our method respectively. It can be seen that our method achieves the highest point cloud accuracy in MAE and RMSE, which in turn reflects the highest accuracy of our LiDAR pose estimation.}
\label{fig:gt-error-map}
\end{figure*}

\begin{table*}[htb!]
\centering
\caption{Evaluation of LiDAR trajectory accuracies of VoxelMap, LOAM and our LiVisSfM by APE and RPE in MAE/RMSE in meters on three KITTI sequences ``00'', ``07'' and ``09''.}
\begin{tabular}{c|clcl|clcl|clcl}
\hline
   & \multicolumn{4}{c|}{VoxelMap} & \multicolumn{4}{c|}{LOAM} & \multicolumn{4}{c}{LiVisSfM} \\
   & \multicolumn{2}{c}{APE} & \multicolumn{2}{c|}{RPE} & \multicolumn{2}{c}{APE} & \multicolumn{2}{c|}{RPE} & \multicolumn{2}{c}{APE} & \multicolumn{2}{c}{RPE} \\ \hline
00 & \multicolumn{2}{c}{23.35/30.60} & \multicolumn{2}{c|}{\textbf{1.15}/\textbf{1.21}} & \multicolumn{2}{c}{6.80/8.33} & \multicolumn{2}{c|}{5.02/10.62} & \multicolumn{2}{c}{\textbf{1.81}/\textbf{2.03}} & \multicolumn{2}{c}{3.09/3.43} \\
07 & \multicolumn{2}{c}{1.23/1.31} & \multicolumn{2}{c|}{1.79/2.00} & \multicolumn{2}{c}{\textbf{0.48/0.52}} & \multicolumn{2}{c|}{2.09/2.82} & \multicolumn{2}{c}{0.49/\textbf{0.52}} & \multicolumn{2}{c}{\textbf{1.73/1.98}} \\
09 & \multicolumn{2}{c}{1.59/2.54} & \multicolumn{2}{c|}{3.00/3.10} & \multicolumn{2}{c}{\textbf{1.49/1.67}} & \multicolumn{2}{c|}{4.05/6.69} & \multicolumn{2}{c}{1.84/1.96} & \multicolumn{2}{c}{\textbf{2.77/2.91}} \\ \hline
\end{tabular}
\label{table:trajectory-error}
\end{table*}

\begin{table*}[htb!]
\caption{Detailed computation time of our LiVisSfM pipeline in all the steps of KITTI sequences ``07'' and ``09'' and cases ``Tearoom'' and ``Corridor'' in minutes.}
\resizebox{\linewidth}{!}{
\begin{tabular}{c|c|c|c|c|c|ccc|c}
\hline
\multirow{2}{*}{} &
  {\# Visual} &
  {\# LiDAR} &
  {Feature} &
  {Feature} &
  {LiDAR \& Visual} &
  \multicolumn{3}{c|}{Mapping} &
  \multirow{2}{*}{Total} \\
         & Frames & Frames & Extraction & Matching & Frame Registration & BA & Loop & Total &  \\ \hline
 07 & 1102 & 551 & 0.401 & 1.543 & 12.0922 & 64.236 & 0.203 & 71.3348 & 85.371 \\
 09 & 1592 & 796 & 0.645 & 3.029 & 20.429 & 85.818 & 0.246 & 93.626 & 117.729 \\
Tearoom & 732 & 733 & 1.675 & 1.283 & 1.8185 & 23.4 & 0 & 30.3485 & 35.125 \\
Corridor & 2788 & 2919 & 12.771 & 5.375 & 4.885 & 60.597 & 0.004 & 76.376 & 99.407 \\ \hline
\end{tabular}
}
\label{table:time-statistics}
\end{table*}

Fig.~\ref{fig:teaser} and \ref{fig:ablation-study} has already shown the reconstruction results of our LiVisSfM on self-captured case ``Outdoor Tianren Office'' and KITTI sequence ``07''. Here, we further give the qualitative comparison of our LiVisSfM to FAST-LIO2~\cite{xu2022fast}, LOAM~\cite{zhang2014loam} and VoxelMap~\cite{yuan2022efficient}
on the self-captured cases ``Corridor'',  ``Teamroom'', and the KITTI sequences ``00'' and ``09'', with the estimated camera trajectories and the fused colorized dense point clouds given in Fig.~\ref{fig:comparison-our-data} and Fig.~\ref{fig:comparison-kitti} respectively. Note that FAST-LIO2 is unable to run on KITTI dataset because it requires a high-frequency IMU input, which is violated by the low-frequency of KITTI dataset, while LOAM cannot handle our self-captured dataset since it doesn't support the data format of MetaCam-Air LiDAR scanner. For KITTI dataset, we sample one frame from every three ones for sequence ``00'', and take one frame for every two ones for ``07'' and ``09'' to improve the computation efficiency. We use pinhole camera model and fisheye distortion model to reconstruct KITTI sequences and self-captured cases respectively to fully adapt to the various image distortion models of the different datasets.

From the recovered camera trajectories and geometric details of the reconstructed dense point clouds, we can see that our LiVisSfM can achieve the best results, especially on self-captured dataset, while FAST-LIO2 shows obvious accumulation errors in the KITTI ``00'' and scale collapsion in case ``Corridor''. VoxelMap has obvious drifts on the self-captured cases because it violates the uniform motion model proposed in~\cite{yuan2022efficient}. The performance of LOAM is relatively stable, but accumulation errors are prone to occur in long-range scenes, as shown for KITTI sequences ``00'' and ``09''. 
In comparison, our LiVisSfM not only achieves comparable reconstruction results to the SOTA methods on KITTI dataset, but also significantly outperforms other approaches on the recovered trajectories and fused LiDAR point clouds for self-captured dataset.

We further provide quantitative evaluation on the accuracies of LiDAR trajectories of our LiVisSfM pipeline and the SOTA methods LOAM~\cite{zhang2014loam} and VoxelMap~\cite{yuan2022efficient} on the KITTI sequences in Table~\ref{table:trajectory-error}, by computing APE and RPE of each method in MAE/RMSE compared to the GT trajectories, which are available from the dataset. It can be seen that our method achieves comparable accuracy in APE and outperforms other methods in RPE. It is particularly worth noting that other methods have obvious accumulation errors in large-scale long-range scenes such as KITTI ``00'', while our method can achieve higher pose accuracy without obvious drift or accumulation error.

Fig.~\ref{fig:gt-error-map} also provides the 
qualitative and quantitative comparison of the fused LiDAR point cloud by our LiVisSfM to SOTA methods FAST-LIO2 and VoxelMap on the self-captured ``ZJU-CCE'' which contains an academic building and a clock tower occupying almost $3000m^2$ area, whose GT 3D model was acquired by laser scanning device for evaluating point-to-model accuracy in both MAE and RMSE. For computing the point cloud accuracy of each method, we use CloudCompare\footnote{http://cloudcompare.org} to compare the reconstructed dense point cloud with GT model in this way: we align the point cloud with GT by manual rough registration followed by an ICP fine registration, and then compute their Point-to-GT-Plane distances for MAE/RMSE evaluation. The whole process can be achieved by CloudCompare’s built-in functions. From the evaluated point cloud accuracies shown in Fig.~\ref{fig:gt-error-map}, we can see that compared to VoxelMap and FAST-LIO2, our LiVisSfM system is able to reconstruct LiDAR point cloud with higher accuracy, which in turn proves the higher reliability of our recovered LiDAR poses.

\subsection{Time Statistics}
Table~\ref{table:time-statistics} gives the time statistics of our pipeline on two KITTI sequences ``07'' and ``09'' and two self-captured cases ``Tearoom'' and ``corridor''. The experiments are conducted on a desktop platform equipped with a $6$-core Intel Xeon i7-8700 CPU @ $3.20$GHz, an Nvidia GeForce RTX 4090 GPU, and $48$GB physical RAM. Both the LiDAR and visual frame registration module and the mapping module run on CPU, while the feature extraction and matching modules are speed up by GPU parrallism. Note that the time consumptions of LiDAR-visual BA module is proportional to the numbers of LiDAR frames, and feature extraction and matching modules are proportional to the number of camera frames.
It is worth noting that the feature extraction and matching modules of self-captured cases consume significantly more time than KITTI sequences, but the LiDAR-visual BA module takes less time. This is because more visual features are extracted from the self-captured cases than KITTI dataset, but the performance frequency of LiDAR-visual BA is lower. Even for the large-scale outdoor scene such as KITTI ``09'', our LiVisSfM can also achieve high-quality 3D point cloud reconstruction in a time-efficient way.

%\subsection{Ablation Studies}

\section{Conclusion and Future Work}
This paper proposes a novel LiDAR-visual SfM framework called LiVisSfM, which first aligns visual and LiDAR maps via ICP in the initialization stage, and alteratively register LiDAR and visual frames to the visual map and LiDAR map in a voxel map representation, by adopting a point-to-Gaussian residual measurement for more accurate poses. Meanwhile, in order to eliminate accumulation errors, we performs LiDAR-visual BA and explicit loop closure, with the voxel map updated in an efficient incremental way to improve the runtime performance of the mapping module. Experiments on KITTI and self-captured dataset demonstrate the reconstruction accuracy of the proposed LiVisSfM system.

Our system is like a one-way LiDAR-visual SfM with visual poses as initial values for LiDAR pose registration. A two-way feedback for joint optimization of visual and LiDAR poses is preferred in the future to better adapt to various complex environments. Besides, dense reconstruction simply fuses LiDAR frames to get a point cloud, with visual frames only for colorization. How to reconstruct a more accurate and complete dense mesh or 3DGS model by fully combining LiDAR points and visual cues remains to be another future work.

\backmatter

\bmhead{Acknowledgments}
The authors wish to thank Chongshan Sheng, Fei Jiao, Qi Chen, Hongliang Sun, Bingze Li, Yuqing Xie, Qiannv Ma, Zhongyun Lu and Huidong Gu for their kind helps in the experiments and the development of the proposed LiVisSfM system, and thank Manhao Situ, Zhichao Feng, Peng Wu, Jie Pan and Shaozu Cao from Skyland Innovation for the development of the MetaCam-Air LiDAR scanner. This work was partially supported by Key R\&D Program of Zhejiang Province (No. 2023C01039).

\section*{Declarations}
\begin{itemize}
\item Availability of data and materials: The datasets generated and analysed during the current study are not publicly available due to privacy protection, but are available from the corresponding author on reasonable request.
\end{itemize}

\bibliography{main}% common bib file

\appendix
\section{Appendix}
\label{sec:appendix}
\subsection{Incremental Voxel Updating}
Suppose a new point ${{\bf{x}}_w}$ is added to voxel ${\bf{v}}$, its updated mean value ${\hat{\mu}_{\bf{v}}}$ is computed as follows:
\begin{equation}
\begin{array}{*{20}{l}}
\because {\mu}_{\bf{v}} = \frac{1}{|\mathcal{L}({\bf{v}})|}{\sum}_{{\bf{x}}_p \in \mathcal{L}({\bf{v}})} {\bf{x}}_p, \\ [6pt]
\therefore {\sum}_{{\bf{x}}_p \in \mathcal{L}({\bf{v}})} {\bf{x}}_p = \mathcal{L}({\bf{v}}) {\mu}_{\bf{v}}
\end{array}
\label{eq:mean-value}
\end{equation}
\begin{equation}
\begin{array}{*{20}{l}}
\hat{\mu}_{\bf{v}} & = \frac{1}{|\mathcal{L}({\bf{v}})| + 1}({\sum}_{{\bf{x}}_p \in \mathcal{L}({\bf{v}})} {\bf{x}}_p + {\bf{x}}_w) \\ [6pt]
                        & = \frac{1}{|\mathcal{L}({\bf{v}})| + 1}(\mathcal{L}({\bf{v}}) {\mu}_{\bf{v}} + {\bf{x}}_w)
\end{array}
\label{eq:appendix-exceptation-add}
\end{equation}

Let $d{\bf{x}}_p(i) = {\bf{x}}_p(i) - {{\mu}_{\bf{v}}}(i)$, ${{\bf{x}}_p(i,j) = {\bf{x}}_p(i) {\bf{x}}_p(j)}$ and ${{\mu_{\bf{v}}}(i,j) = {\mu_{\bf{v}}}(i) {\mu_{\bf{v}}}(j)}$. The updated covariance matrix ${\hat{\Sigma}_{\bf{v}}}$ is computed as:
\begin{equation}
\begin{array}{*{20}{l}}
\because {{\Sigma}_{\bf{v}}}(i,j) = \frac{1}{|\mathcal{L}({\bf{v}})|} \sum_{{\bf{x}}_p \in \mathcal{L}({\bf{v}})} d{\bf{x}}_p(i) d{\bf{x}}_p(j) \\ [6pt]
\quad \quad \quad \quad = \frac{1}{|\mathcal{L}({\bf{v}})|} \sum_{{\bf{x}}_p \in \mathcal{L}({\bf{v}})} {\bf{x}}_p(i,j) - {\mu_{\bf{v}}}(i,j) \\ [6pt]
\therefore \sum_{{\bf{x}}_p \in \mathcal{L}({\bf{v}})} {\bf{x}}_p(i,j) = |\mathcal{L}({\bf{v}})|({\Sigma_{\bf{v}}}(i,j) + {\mu_{\bf{v}}}(i,j)) \\
\end{array}
\label{eq:covariance}
\end{equation}
\begin{equation}
\begin{array}{*{20}{l}}
{\hat{\Sigma}_{\bf{v}}}(i,j) \\ [6pt]
= \frac{1}{|\mathcal{L}({\bf{v}})|+1} (\sum_{{\bf{x}}_p \in \mathcal{L}({\bf{v}})} {\bf{x}}_p(i,j) + {\bf{x}}_w(i,j)) \\ [6pt]
\quad - {\hat{\mu}_{\bf{v}}}(i) {\hat{\mu}_{\bf{v}}}(j) \\ [6pt]
= \frac{1}{|\mathcal{L}({\bf{v}})|+1} (|\mathcal{L}({\bf{v}})|({\Sigma_{\bf{v}}}(i,j) + {\mu_{\bf{v}}}(i,j)) + {\bf{x}}_w(i,j)) \\ [6pt]
\quad - {\hat{\mu}_{\bf{v}}}(i) {\hat{\mu}_{\bf{v}}}(j) \\ [6pt]
= \frac{|\mathcal{L}({\bf{v}})|({\Sigma_{\bf{v}}}(i,j) + {\mu}_{\bf{v}}(i) {\mu}_{\bf{v}}(j)) + {\bf{x}}_w(i) {\bf{x}}_w(j)}{|\mathcal{L}({\bf{v}})|+1} - {\hat{\mu}_{\bf{v}}}(i) {\hat{\mu}_{\bf{v}}}(j),
\end{array}
\label{eq:appendix-covariance-add}
\end{equation}
which is exactly the same as Eq.~\ref{eq:mean-covariance-add}.

Similarly, when point ${\bf{x}}_w$ is removed from voxel ${\bf{v}}$, the mean $\hat{\mu}_{\bf{v}}$ and covariance $\hat{\Sigma}_{\bf{v}}$ can be expressed as:
\begin{equation}
\begin{array}{*{20}{l}}
\hat{\mu}_{\bf{v}} & = \frac{1}{|\mathcal{L}({\bf{v}})| - 1}({\sum}_{{\bf{x}}_p \in \mathcal{L}({\bf{v}})} {\bf{x}}_p - {\bf{x}}_w) \\ [6pt]
                        & = \frac{1}{|\mathcal{L}({\bf{v}})| - 1}(\mathcal{L}({\bf{v}}) {\mu}_{\bf{v}} - {\bf{x}}_w)
\end{array}
\label{eq:appendix-exceptation-sub}
\end{equation}

\begin{equation}
\begin{array}{*{20}{l}}
{\hat{\Sigma}_{\bf{v}}}(i,j) \\ [6pt]
= \frac{1}{|\mathcal{L}({\bf{v}})| - 1} (\sum_{{\bf{x}}_p \in \mathcal{L}({\bf{v}})} {\bf{x}}_p(i,j) - {\bf{x}}_w(i,j)) \\ [6pt]
\quad - {\hat{\mu}_{\bf{v}}}(i) {\hat{\mu}_{\bf{v}}}(j) \\ [6pt]
= \frac{1}{|\mathcal{L}({\bf{v}})| - 1} (|\mathcal{L}({\bf{v}})|({\Sigma_{\bf{v}}}(i,j) + {\mu_{\bf{v}}}(i,j)) - {\bf{x}}_w(i,j)) \\ [6pt]
\quad - {\hat{\mu}_{\bf{v}}}(i) {\hat{\mu}_{\bf{v}}}(j) \\ [6pt]
= \frac{|\mathcal{L}({\bf{v}})|({\Sigma_{\bf{v}}}(i,j) + {\mu}_{\bf{v}}(i) {\mu}_{\bf{v}}(j)) - {\bf{x}}_w(i) {\bf{x}}_w(j)}{|\mathcal{L}({\bf{v}})| - 1} - {\hat{\mu}_{\bf{v}}}(i) {\hat{\mu}_{\bf{v}}}(j),
\end{array}
\label{eq:appendix-covariance-sub}
\end{equation}
which is the same as Eq.~\ref{eq:mean-covariance-sub}.

\end{document}